%% file: main.tex
\documentclass[sigconf, nonacm]{acmart}

\AtBeginDocument{%
  }

\usepackage{xspace}
\usepackage{times}
\usepackage{latexsym}
\usepackage{amsmath}
\usepackage{multicol}
\usepackage{multirow}
\usepackage{graphicx}
\usepackage{longtable}
\usepackage{tabularx}
\usepackage{balance}
\usepackage{booktabs}
\usepackage{tcolorbox}
\usepackage{xcolor}
\usepackage{float}
\usepackage{enumitem}
\usepackage{verbatim}
\usepackage{listings}
\usepackage{thm-restate}
\usepackage{mathtools}

\begin{document}
\title{Token-level Proximal Policy Optimization for Query Generation}

\author{Yichen Ouyang\textsuperscript{1,†}, 
Lu Wang\textsuperscript{2,*}, 
Fangkai Yang\textsuperscript{2}, 
Pu Zhao\textsuperscript{2}, Chenghua Huang\textsuperscript{3,†}, 
Jianfeng Liu\textsuperscript{2}, 
Bochen Pang\textsuperscript{2}, Yaming Yang\textsuperscript{2}, 
Yuefeng Zhan\textsuperscript{2}, 
Hao Sun\textsuperscript{2}, Qingwei Lin\textsuperscript{2}, 
Saravan Rajmohan\textsuperscript{2}, 
Weiwei Deng\textsuperscript{2}, Dongmei Zhang\textsuperscript{2}, 
Feng Sun\textsuperscript{2}, 
Qi Zhang\textsuperscript{2}}

\affiliation{
\begin{minipage}{\linewidth}
\centering
\textsuperscript{1}Zhejiang University \quad
\textsuperscript{2}Microsoft \quad
\textsuperscript{3}Fudan University \\
\textsuperscript{1}22271110@zju.edu.cn \quad \textsuperscript{*}wlu@microsoft.com
\country{}
\end{minipage}
}

\renewcommand{\shortauthors}{Trovato et al.}

\input{tex/abstract}

\maketitle

\input{tex/introduction}
\input{tex/related_work}
\input{tex/method}
\input{tex/experiment}
\input{tex/ablation}
\input{tex/conclusion}

\bibliographystyle{ACM-Reference-Format}
\bibliography{ref}
\input{tex/appendix}

\end{document}

%% file: tex/abstract.tex
\begin{abstract}

Query generation is a critical task for web search engines (e.g. Google, Bing) and recommendation systems. Recently, state-of-the-art query generation methods leverage Large Language Models (LLMs) for their strong capabilities in context understanding and text generation. However, they still face challenges in generating high-quality queries in terms of inferring user intent based on their web search interaction history. In this paper, we propose Token-level Proximal Policy Optimization (TPPO), a noval approach designed to empower LLMs perform better in query generation through fine-tuning. TPPO is based on the Reinforcement Learning from AI Feedback (RLAIF) paradigm, consisting of a token-level reward model and a token-level proximal policy optimization module to address the sparse reward challenge in traditional RLAIF frameworks. To evaluate the effectiveness and robustness of TPPO, we conducted experiments on both open-source dataset and an industrial dataset that was collected from a globally-used search engine. The experimental results demonstrate that TPPO significantly improves the performance of query generation for LLMs and outperforms its existing competitors.

\end{abstract}

%% file: tex/introduction.tex
\section{Introduction}

\makeatletter
\newcommand\footnotetextstar[2]{%
    \renewcommand\@thefnmark{#1}%
    \footnotetext{#2}}
\makeatother

\renewcommand{\thefootnote}{\fnsymbol{footnote}}
\footnotetext[1]{Corresponding author.}  
\footnotetext[2]{Work is done during an internship at Microsoft.}

\renewcommand{\thefootnote}{\arabic{footnote}}
\setcounter{footnote}{0} 
\footnotetext[1]{Throughout the paper, we use the term ``sentence-level'' to represent the sparse reward cases where reward is given at the end of a response or each sentence.}

Web query generation is an essential ingredient for search engines~\cite{he2009web,aggarwal2016recommender,cai2016survey,wu2018query}. A common issue in web search is that the user often needs multiple iterations of query refinement in order to get desired results. The task of web query generation is to make the generated queries align with users' personal preferences that better represent their search intent during web search. Such personalized web query is inferred from user's historical search records and should be relevant and meaningful to each user~\cite{baek_knowledge-augmented_2024,yang2023palrpersonalizationawarellms}. It is particularly important for the current personalized search engines such as Bing and Google. The advent of Large Language Models (LLMs) has achieved superior performance in search engines and recommendation systems, relying on their ability to understand textual features and generate high-quality responses~\cite{li2023gpt4rec,zhao2023recommender,wu2024survey}. Despite the great success of LLMs, there still exist challenges in domain-specific tasks such as web query generation in terms of inferring user intents from historical short and ambiguous search queries and generating relevant while diverse web search queries. 

Reinforcement learning from external feedback such as Reinforcement Learning from Human Feedback (RLHF)~\cite{christiano2017deep,ziegler2019fine,stiennon2020learning,ouyang2022training,bai2022training} and Reinforcement Learning from AI Feedback (RLAIF)~\cite{bai2022constitutional,lee2023rlaif} is a mainstream approach to aligning LLMs with human values and preferences. It introduces human feedback into Reinforcement Learning (RL), learning reward functions and optimizing policies to enable LLMs to generate helpful responses aligned with human~\cite{ouyang2022training, ziegler2019fine}. This approach makes LLMs better adapt to domain-specific tasks to improve performance~\cite{kirk2023understanding,wang2023survey,ge2024openagi}. In this paper, we are the first to explore the application of RLAIF on the web query generation task to optimize LLMs in generating personalized search queries aligned with each user's preference and interest. 

\begin{figure}[t]
\centering
\includegraphics[width=0.5\textwidth]{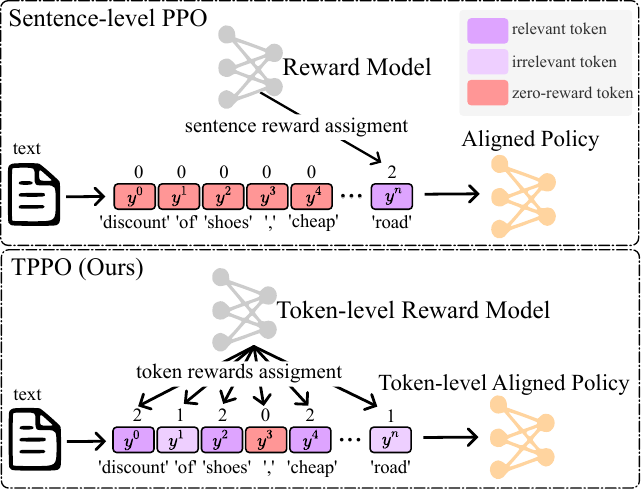}
\caption{
Reward assignment in sentence-level PPO and token-level PPO (TPPO). Sentence-level PPO assigns reward only at the end of a response, whereas TPPO assigns reward for each token in a response.
}
\label{fig:token-level ppo}
\end{figure}

\begin{figure*}[h]
\centering
\includegraphics[width=1\textwidth]{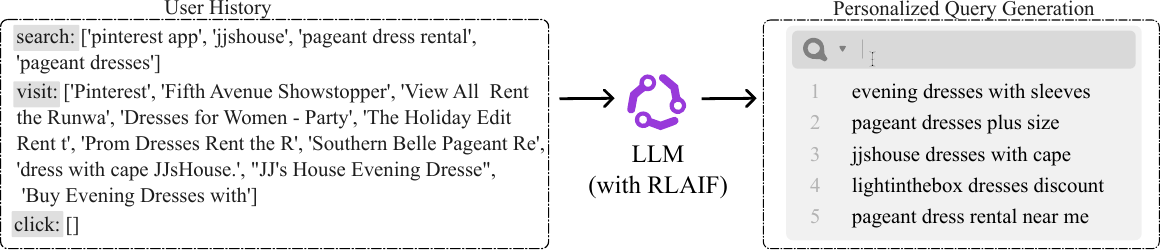}
\caption{The Query Generation Task. Taking user history as input, the LLM after RLAIF alignment outputs several personalized queries that the user is interested in.}
\label{fig:query gen task}
\end{figure*}

The process of RLAIF consists of two steps: the reward model training from AI generated feedback and the reward-based RL training that optimizes the LLM's performance in response generation. As a seminal policy gradient algorithm in RL, Proximal Policy Optimization (PPO)~\cite{schulman2017proximalpolicyoptimizationalgorithms} plays a key role in optimizing agent policies within the RLAIF framework. However, the training of PPO is known to be unstable~\cite{christiano2017deep,rafailov2024direct,zhong2024dpo} and one potential reason could be that the reward signal is typically provided for the entire response, i.e., at the end of the response sentence, making the reward sparse. The sparse reward makes current PPO in RLAIF actually be sentence-level\footnotemark[1]. The sentence-level PPO has some inherent limitations and poses difficulties. 
Firstly, the sparsity of sentence-level rewards poses challenges~\cite{guo2024imitationleveragingfinegrainedquality, wu2023finegrainedhumanfeedbackgives, rafailov2024directpreferenceoptimizationlanguage}. The reward is provided at the end of a sentence or a response, where the entire sentence is considered to be an action. However, a sentence consists of tokens and the generation of each token is an action. Many intermediate actions receive no explicit feedback. This sparsity can lead to inefficient exploration and causes sentence-level PPO to struggle in finding the optimal policy, and it makes the trained policy hard to distinguish between good and bad actions within a sentence as the reward signal is not fine-grained enough to provide meaningful feedback at the token level. Moreover, the sentence-level PPO suffers from the temporal delay~\cite{arjonamedina2019rudderreturndecompositiondelayed, hung2018optimizingagentbehaviorlong} between the actions (token generation) and the reward signal, introducing training instability and divergence in policy training since many actions (token generations) occur before receiving any feedback.
Secondly, the mismatch between traditional PPO's formulation in RL and the sentence-level reward signal introduces difficulties for sentence-level PPO~\cite{uesato2022solving, lightman2023let}. PPO is designed for multi-step RL training where the value function is typically learned to estimate the expected future rewards at each step. However, with sentence-level rewards, the value function could not accurately capture the long-term impact of individual actions, leading to sub-optimal policy updates.

To address above limitations and challenges, our work proposes a token-level PPO (TPPO) as shown in Figure~\ref{fig:token-level ppo}. Specifically, by leveraging token-level reward models and corresponding token-level PPO policies, we mitigate the issues of sparse rewards and design a token-level PPO which matches with the traditional PPO in RL, increasing the training stability.
Firstly, to tackle the sparsity of sentence-level rewards and provide more fine-grained feedback, we propose a token-level reward model for RLAIF to assign rewards to individual tokens or actions within a sentence. 
Secondly, to address the mismatch between PPO's formulation and the sentence-level reward signal, we introduce a token-level PPO policy. This token-level PPO policy is designed to align with the token-level reward model. By learning a value function that estimates the expected future rewards at the token level, the policy can make more informed decisions based on the immediate impact of each action. 
This alignment between the token-level reward model and the token-level PPO policy helps mitigate the sub-optimal policy updates. Besides, by assigning rewards to individual tokens, the algorithm can more accurately assign credit to specific actions, leading to more stable and consistent updates. 

We conduct experiments on both industrial dataset and public benchmarks. The results show that TPPO can increase the overall relevance rate of query generation by 2\%-4\% compared to PPO, and has a higher win rate than PPO at 2\%-8\% in item-by-item comparisons. The training process reveals that TPPO exhibits better convergence behavior, with a steadily increasing reward, smaller variance, and improved loss, highlighting its superior preference alignment capability. Notably, our model has been successfully deployed in real-world applications, further validating its practical effectiveness and scalability. Our key contributions are summarized as follows:
\begin{itemize}
    \item We propose token-level Proximal Policy Optimization (TPPO) for RLAIF, incorporating key components such as token-level reward labeling, reward model training, and token-level Proximal Policy Optimization.
    \item We are the first to adopt TPPO to empower the query generation task which will benefit both academia and industry.
    \item Comprehensive experiments on industrial datasets and public benchmarks validate the effectiveness, practicality, and robustness of our approach.
\end{itemize}

%% file: tex/related_work.tex
\section{Related Work} 

\subsection{Query Generation}
Query generation in web search is the process of generating new search queries that align with a user's interests and preferences, based on their previous search history, browsing behaviors, and other contextual information. The goal is to anticipate the user's future information needs and provide them with relevant search suggestions or recommendations~\cite{he2009web,aggarwal2016recommender,cai2016survey}. 
However, query generation faces challenges such as accurately inferring user intent from short and ambiguous queries, understanding the context of the user's search session, and balancing the trade-off between relevance and diversity in generated queries~\cite{mustar2021study,jannach2022session}. Recent works leverage LLMs in terms of query generation or query suggestion for recommendation systems~\cite{li2023gpt4rec,zhao2023recommender,wu2024survey,wei2024llmrec,lin2024recommendersystemsbenefitlarge,li2024surveygenerativesearchrecommendation}. For example, GPT4Rec~\cite{li2023gpt4rec} leverage generated search queries to retrieve items for recommendation by searching these queries. \citet{baek_knowledge-augmented_2024} leverage users' interaction histories to personalize the search queries. Despite the rich knowledge and strong in-context learning capabilities of LLMs, the performance of LLMs in domain-specific tasks still remian suboptimal due to the substantial difference between the training tasks and the domain-specific tasks such as query generation tasks, as well as inadequate domain knowledge in the model pretraining~\cite{bao2023tallrec,zhang2023reaugkd,yang2023empower,wang2024can}. Reinforcement Learning from AI Feedback (RLAIF)~\cite{bai2022constitutional,lee2023rlaif} is a mainstream approach to aligning LLMs with human values and preferences, demonstrating better performance in domain-specific tasks. In this paper, we are the first to leverage RLAIF in addressing the aforementioned challenges in the query generation task, allowing the trained LLM to generate queries that better align with user preferences. In particular, we creatively propose token-level reward and training policy, help LLMs stablize the training and capture the nuances and user intent more effectively.

\subsection{Proximal Policy Optimization}

Proximal Policy Optimization (PPO) ~\cite{schulman2017proximalpolicyoptimizationalgorithms} is a popular and effective algorithm for policy optimization in reinforcement learning. It improves traditional policy gradient methods with an objective function that prevents large updates, thus ensuring more stable and reliable training~\cite{Kakade2002ApproximatelyOA}. Recently, researchers have explored the usage of PPO in the context of RLAIF for natural language processing (NLP) tasks~\cite{ziegler2019fine, bai2022training,yue2023llamarectwostagerecommendationusing}. However, adapting PPO in RLAIF suffers from unstable training~\cite{rafailov2024direct,zhong2024dpo}. Unlike traditional PPO where the reward is given to each action, the PPO in current RLAIF considers the entire response as an action and gives reward at the end of each response. In practice, LLM generates response in a next-token prediction manner where each token is an action. The sentence-level reward which gives at the end of the response is a sparse reward, making the PPO actual sentence-level. The mismatch between traditional PPO and the sentence-level PPO in RLAIF framework leads to issues such as inefficient exploration, sub-optimal policy updates, and training instability~\cite{xia_inverse-q_nodate, xu_aligning_2024}. In this paper, we propose token-level PPO which gives reward to each token in a response to mitigate the sparse reward and temporal delay issues~\cite{arjonamedina2019rudderreturndecompositiondelayed, hung2018optimizingagentbehaviorlong}, matching the PPO in RLAIF framework with the traditional PPO in RL, improving the training stability and enhancing the performance of trained LLMs in the domain-specific query generation task in web search.

%% file: tex/method.tex
\section{Methodology}

\begin{figure*}[h]
\centering
\includegraphics[width=0.9\textwidth]{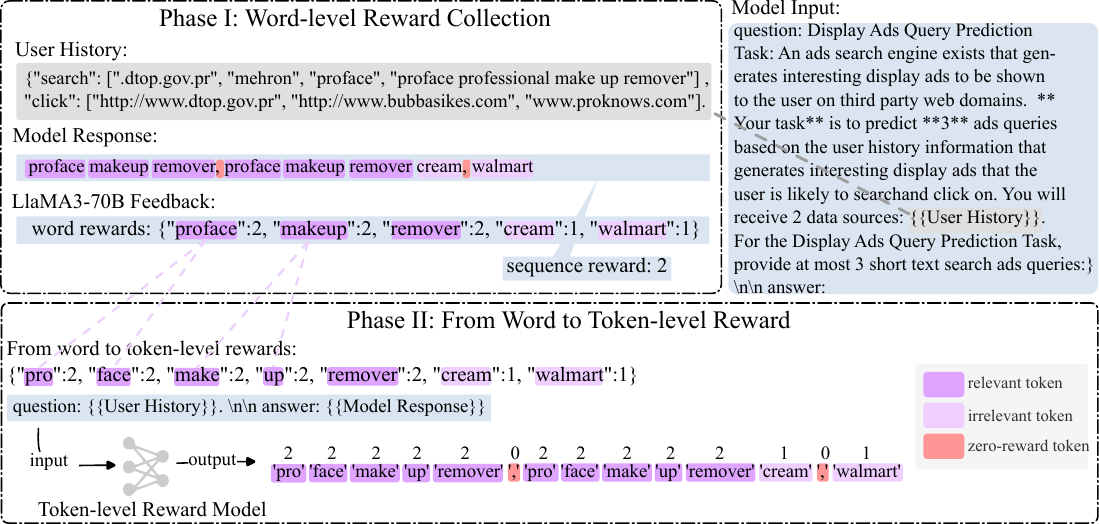}
\caption{Token-level reward labeling. In phase I, we use LLaMA 3 (70B) to label word-level and sentence-level rewards for the dataset. In phase II, we map the word-level rewards to token-level rewards. The model response and user history are used to construct input for token-level reward model and the mapped token rewards are used as the ground truth for output.}
\label{fig:token-level rm}
\end{figure*}

In this section, we introduce the problem formulation for the query generation task in Section~\ref{sec:problemformulation} and we then describe the workflow of our token-level PPO within RLAIF framework consists of token-level reward labeling (Section~\ref{sec:rewardlabeling}), reward model training (Section~\ref{sec:rewardmodeltraining}), and LLM training with token-level PPO (Section~\ref{sec:llmtraining}).

\subsection{Problem Formulation}\label{sec:problemformulation}
We formulate the query generation task as a sequential token generation problem. Given an input prompt $\mathbf{x}$ and the previously generated $t-1$ tokens $\{y^{<t}\}=[y^1, y^2, \dots, y^{t-1}]$ of the query\footnote{The initial token is generated given the prompt $\mathbf{x}$ only.}, the language model, i.e., the policy $\pi_{\theta}$ predicts the probability distribution of the next token $\pi_{\theta}(\cdot|\mathbf{x}, \{y^{<t}\})$. 
In the our token-level PPO formulation, the \emph{state} of the $t^{\text{th}}$ step $s_t$ is a concatenation of the input prompt and the generated response up to this step, denoted as $s_t = [\mathbf{x}, \{y^{<t}\}]$. An \emph{action} corresponds to the next generated token, denoted as $a_t = y^t$, and the \emph{reward} at this step is defined as $R_t = R(s_t, a_t)$. 
Our objective is to maximize the expected cumulative reward over the sequence of tokens generated by the policy $\pi_{\theta}$.
The state-action value function is defined as:
$Q_{\pi_{\theta}}(s_t, a_t) = \sum_{k=0}^{\infty}\gamma^{k} R_{t+k}$.
Then, we define the state value function $V_{\pi_{\theta}}(s_t, a_t)  = \mathbb{E}_{a_t \sim \pi_{\theta}}\left[ Q_{\pi_{\theta}}(s_t, a_t)\right]$ and the advantage function $A_{\pi_{\theta}}(s_t, a_t)  = Q_{\pi_{\theta}}(s_t, a_t) - V_{\pi_{\theta}}(s_t, a_t)$ for a policy $\pi_{\theta}$.

\subsection{Token-Level Reward Labeling}\label{sec:rewardlabeling}

Labeling token-level rewards with human annotation is expensive and time-consuming, and LLM-based annotation achieves comparable performance as human labeling\cite{bai2022constitutional,lee2023rlaif,zheng2023judging,chen2024humans}. In this paper, we adopt LLaMA 3 (70B) to label token-level rewards due to its strong labeling capability~\cite{touvron2023llama} (Phase I in Figure~\ref{fig:token-level rm}). Compared with sentence-level reward which overlooks the impact of individual tokens~\cite{zeng2024token,cao2024drlc,zeng2024token}, the token-level reward which is assigned to each token, capturing finer-grained feedback. On the other hand, the sentence-level reward provides a holistic feedback on the entire generated query/response, and we include the sentence-level reward as a constraint to guide the token-level rewards as it is typically easier and less prone to noise.  

We design annotation prompts to instruct LLaMA 3 to score the relevance of each token within the generated query and the user’s history (token-level reward), followed by an overall relevance score for the entire query (sentence-level reward). Each reward is assigned to one of the labeling categories consist of 0, 1, or 2, where 0 represents non-reward tokens or masked tokens (e.g., comma separators), 1 represents irrelevant tokens, and 2 represents relevant tokens. Only rewards of irrelevant and relevant tokens are used in updating the PPO policy. Then we collect both token-level and sentence-level rewards after reward labeling.

\noindent\textbf{From word-level reward to token-level reward.} Labeling at the word-level is more manageable for LLaMA 3 compared to token-level annotation, as different models use different tokenizers. Labeling at the word level makes the rewards become more generalizable across models, even when they employ distinct tokenization methods. We then map the labeled word-level rewards in Phase I of Figure~\ref{fig:token-level rm} to token-level rewards. For example, the word \texttt{relevant} is assigned with a word-level reward of category 2. Depending on the tokenizer, the word could split into:
\begin{itemize}
    \item Model 1: \texttt{["re", "levant"]}
    \item Model 2: \texttt{["relev", "ant"]}
\end{itemize}
We assign the same reward category 2 to each token:
\begin{itemize}
    \item Model 1: \texttt{"re" $\rightarrow$ 2}, \texttt{"levant" $\rightarrow$ 2}
    \item Model 2: \texttt{"relev" $\rightarrow$ 2}, \texttt{"ant" $\rightarrow$ 2}
\end{itemize}

\begin{figure*}[t]
\centering
\includegraphics[width=0.9\textwidth]{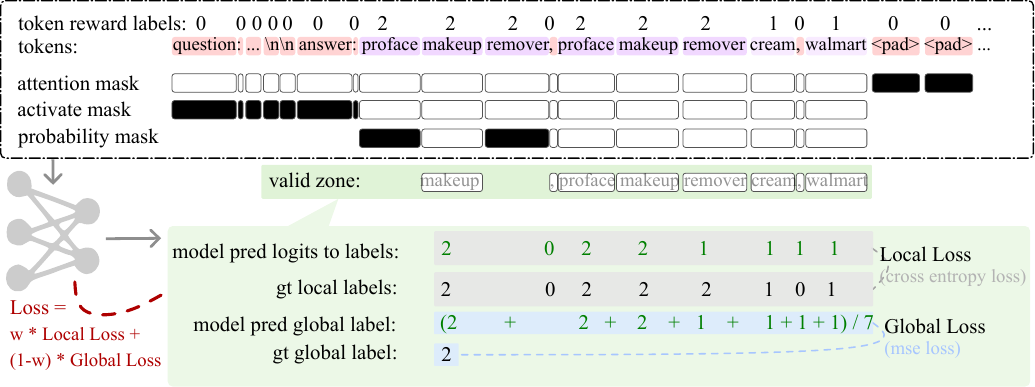}
\caption{Objectives of token-level reward model. The position after masking (valid zone) is used to calculate the loss and return gradient. The loss of the token-level reward model is the weighted sum of local loss and global loss.}
\label{fig:objectives}
\end{figure*}

\subsection{Reward Model Training}\label{sec:rewardmodeltraining}

As shown in Figure~\ref{fig:objectives}, using the token-level and sentence-level rewards labeled by LLaMA 3, we design a local loss (based on token-level rewards) and a global loss (based on sentence-level rewards) for training the token-level reward model.

\noindent\textbf{Local loss of reward model.}  
First, we apply attention and activation masks to exclude padding areas and non-model response tokens. To ensure balanced class representation, we employ a probability mask to maintain the ratio of label 2 to label 1 within the range of approximately 1:3 to 3:1, while preserving tokens with label 0.

After applying these masks, the remaining tokens form the \emph{valid set}, denoted as \( \mathcal{V} \), which is used for loss computation and gradient back-propagation. The local loss is defined as a weighted cross-entropy loss over a batch of \(n\) samples, predicting the probability of each token belonging to one of the three reward categories \( \{0, 1, 2\} \). The loss function is expressed as:
{\small
\begin{equation}
\mathcal{L}_{\text{local}}(\phi) = - \frac{1}{n} \sum_{i=1}^{n} \sum_{(s_t, a_t) \in \mathcal{V}} \sum_{c=0}^{2} w_c \, \mathbf{1}_{[R_{\phi}(s_t, a_t) = c]} \log P(c \mid s_t, a_t),
\end{equation}
}

where \(P(c \mid s_t, a_t)\) is the predicted probability that token \((s_t, a_t)\) belongs to class \(c\) given by the reward model $R_{\phi}$ parameterized by \(\phi\), and \(w_c\) is a class weight. The indicator function \( \mathbf{1}_{[\cdot]} \) equals 1 when the condition holds, and 0 otherwise.

\noindent\textbf{Global loss of reward model.}  
The global loss provides partial supervision by aligning the average token rewards in \( \mathcal{V} \) with the sentence-level reward. It is formulated as the mean squared error (MSE) loss over \(n\) samples, measuring the difference between the average token reward and the ground truth global reward:

\begin{equation}
\mathcal{L}_{\text{global}}(\phi) = \frac{1}{n} \sum_{i=1}^{n} \left( \frac{1}{|\mathcal{V}|} \sum_{(s_t, a_t) \in \mathcal{V}} R_{\phi}(s_t, a_t) - R_{\text{global}} \right)^2,
\end{equation}

where \( |\mathcal{V}| \) is the number of tokens in the valid set, \(R_{\phi}(s_t, a_t) = \arg \max_{c \in \{0, 1, 2\}} P(c \mid s_t, a_t) \) is the predicted token reward, and \( R_{\text{global}} \) is the ground truth global reward. This loss encourages consistency between token-level and sentence-level rewards, ensuring coherent supervision across different levels of granularity.

The total loss for training the reward model combines the local and global losses:

\begin{equation}
\mathcal{L}_{\text{total}}(\phi) = \lambda_{\text{local}} \mathcal{L}_{\text{local}}(\phi) + \lambda_{\text{global}} \mathcal{L}_{\text{global}}(\phi),
\end{equation}

where $\lambda_{\text{local}}$ and $\lambda_{\text{global}}$ are hyperparameters controlling the trade-off between local and global supervision.

\noindent\textbf{Length-weighted penalty.}
When applying the trained token reward model with PPO, to prevent the LLM model from generating overly long responses, we introduce a length-weighted penalty ($lwp$) mechanism to adjust the rewards. The $lwp$ function is defined as:
\begin{equation}
lwp(l) = \frac{1}{1 + e^{\alpha (l - sl) - 6}},
\label{eqn:lengthpenalty}
\end{equation}

where $l$ represents the position of the current token in the model response, $sl$ is the suggested length (an estimated reasonable response length), and $\alpha$ is a hyperparameter controlling the penalty intensity.
Note that Equation~\ref{eqn:lengthpenalty} is designed with the purpose that for tokens with positions before the suggested length ($l \leq sl$), their corresponding $lwp$ equals approximately 1, while for tokens with positions after the suggested length ($l > sl$), their corresponding $lwp$ rapidly decreases towards 0. We multiply the original token-level rewards by the length penalty at the corresponding positions to modify the rewards:

\begin{equation}
R'_{\phi}(s_t, a_t) = lwp(l) \cdot R_{\phi}(s_t, a_t).
\end{equation}

\subsection{LLM Training with Token-Level PPO}\label{sec:llmtraining}

We introduce token-level PPO where the formulation is matched with token-level reward signal. 

\noindent\textbf{Token-level PPO objective function.}
The token-level PPO objective function is formulated as:
\begin{equation}
\begin{split}
\max_{\pi_{\theta}} & \ \mathbb{E}_{\mathbf{x}, y^{<t} \sim \mathcal{D},\, y^t \sim \pi_{\theta}(\cdot \mid [\mathbf{x}, y^{<t}])} \bigg[ \min \left( r_t(\theta) A_{\pi_{\mathrm{ref}}}(s_t, a_t), \right. \\
& \left. \text{clip}\left( r_t(\theta), 1 - \epsilon, 1 + \epsilon \right) A_{\pi_{\mathrm{ref}}}(s_t, a_t) \right) \bigg],
\end{split}
\label{ppo_obj}
\end{equation}

where 
\begin{equation}
r_t(\theta) = \frac{\pi_{\theta}(a_t \mid s_t)}{\pi_{\mathrm{ref}}(a_t \mid s_t)}
\end{equation}
is the ratio between the new policy \( \pi_{\theta} \) and the old policy \( \pi_{\mathrm{ref}} \) at the token level. Here, \( \epsilon \) is a hyperparameter controlling the clipping range, and \( A_{\pi_{\mathrm{ref}}}(s_t, a_t) \) is the advantage function based on the reference policy \( \pi_{\mathrm{ref}} \).

\noindent\textbf{Derivation of the optimal policy.}
Starting from the token-level PPO objective in Equation~\ref{ppo_obj}, we aim to derive the optimal policy \( \pi_{\theta}^* \). To ensure the policy remains close to a reference policy \( \pi_{\mathrm{ref}} \), we introduce a Kullback–Leibler (KL) divergence constraint. The optimization problem is formulated as:

\begin{multline}
\pi_{\theta}^* = \arg\max_{\pi_{\theta}} 
\mathbb{E}_{s_t \sim \mathcal{D},\, a_t \sim \pi_{\theta}(\cdot \mid s_t)} 
\Big[ A_{\pi_{\mathrm{ref}}}(s_t, a_t)  \\
- \beta \, \mathrm{KL}\left( \pi_{\theta}(\cdot \mid s_t) \,\|\, 
\pi_{\mathrm{ref}}(\cdot \mid s_t) \right) \Big],
\label{eq:optimization}
\end{multline}

where \( \beta > 0 \) controls the strength of the KL divergence regularization. The closed-form solution to the optimization problem in Equation~\ref{eq:optimization} is:

\begin{equation}
\pi_{\theta}^*(a_t \mid s_t) = \frac{\pi_{\mathrm{ref}}(a_t \mid s_t) \exp\left( \frac{1}{\beta} A_{\pi_{\mathrm{ref}}}(s_t, a_t) \right)}{Z(s_t; \beta)},
\label{eq_3}
\end{equation}

where \( Z(s_t; \beta) = \sum_{a_t} \pi_{\mathrm{ref}}(a_t \mid s_t) \exp\left( \frac{1}{\beta} A_{\pi_{\mathrm{ref}}}(s_t, a_t) \right) \) is the partition function ensuring that \( \pi_{\theta}^* \) is a valid probability distribution.

\begin{lemma}
\label{lemma4}
The optimization problem in Equation~\ref{eq:optimization} yields the optimal policy as given in Equation~\ref{eq_3}.
\end{lemma}

%% file: tex/experiment.tex
\section{Experiments}

To evaluate the effectiveness of our method, we did experiments on widely used open-source benchmark. Moreover, to evaluation the robustness and practicability of our method, we did experiments on one industrial dataset which collect from a popular search engine that serve billions of worldwide customers.

In the experimental results, we evaluate the performance of each model on the query generation task using two metrics. (1) \textbf{Relevance rate}: The relevance score \( \in \{0, 1\} \) reflects the alignment between the generated query and the input user history, obtained by prompting LLaMA-70B.  We further calculate the \textbf{Relevance rate} by dividing the sum of the relevance scores by the total number of samples in the evaluation set.
 (2) \textbf{Win-Tie-Lose rates}: which is derived from pairwise comparisons of relevance scores across different models. These comparisons are conducted by prompting LLaMA-70B to vote on which model performs better for each data point. Furthermore we analyze the training curves to understand the learning dynamics and convergence behavior of the models. The evaluation prompts used to obtain the relevance scores is provided in the Appendix B.  

\subsection{Experiments on Open-source Data}

\subsubsection{Dataset Description}

The AOL dataset is an open collection consisting of approximately 20 million web queries collected from about 650k users over three months ~\cite{macavaney2022reproducingpersonalisedsessionsearch}. This collection provides real query log data based on real users, which can be used for personalization, query reformulation, or other types of search research. As shown in Figure~\ref{tab:dataset_info}, in our work, we filtered 27k data from AOL dataset to construct an open-source dataset for the query generation task. Specifically, for each data point in our constructed dataset, the user history consists of earlier search queries and click records from the AOL dataset, while the newest 3 search queries are used as the target queries to be generated. For supervised finetuning, 25k data is used for training and 2k for evaluation. For token-level reward model training, we use 10k train data labeled with Llama3-70B. For token-level ppo training, we use 20k train data for training and the full 2k evaluation data for evaluating.

\begin{table}[htbp]
\caption{Dataset information.}
\centering
\scalebox{0.8}{
\begin{tabular}{|l|c|c|}
\hline
 & \textbf{Industrial Dataset} & \textbf{Open Dataset} \\
\hline
\textbf{User History Keys} & 'search', 'click', 'purchase', 'visit' & 'search', 'click' \\
\hline
\multirow{2}{*}{\textbf{Dataset Size}} & Train: 200k, & Train: 27k, \\
 & Eval1: 2k, Eval2: 2k & Eval: 2k \\
\hline
\textbf{Tagert Query Num} & 10 & 3 \\
\hline
\end{tabular}
}
\label{tab:dataset_info}
\end{table}

\subsubsection{Results of Token-level PPO Policy}

\begin{figure}[h!]
\centering
\includegraphics[width=0.48\textwidth]{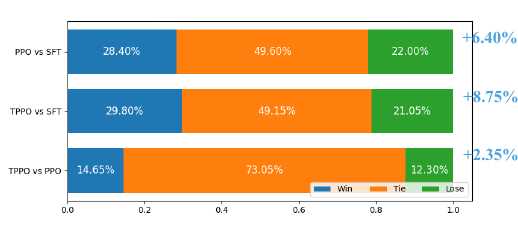}
\caption{Win-Tie-Lose Comparisons on Open-source Dataset.}
\label{fig:gsb_aol_tp1}
\end{figure}

As illustrated in Figure~\ref{fig:gsb_aol_tp1} and Table~\ref{tab:aol_t1}, we compared the relevance scores of models obtained using SFT, PPO, and our TPPO on the same evaluation set, using two methods: overall average scores and pairwise win-lose comparisons. The results demonstrate that TPPO's preference fitting ability consistently outperforms both SFT and PPO. 
Specifically, in the win-tie-lose comparison for relevance rate, the win rate of TPPO is 8.75\% and 2.35\% higher than that of SFT and PPO respectively.

\begin{figure}[t!]
\centering
\includegraphics[width=0.5\textwidth]{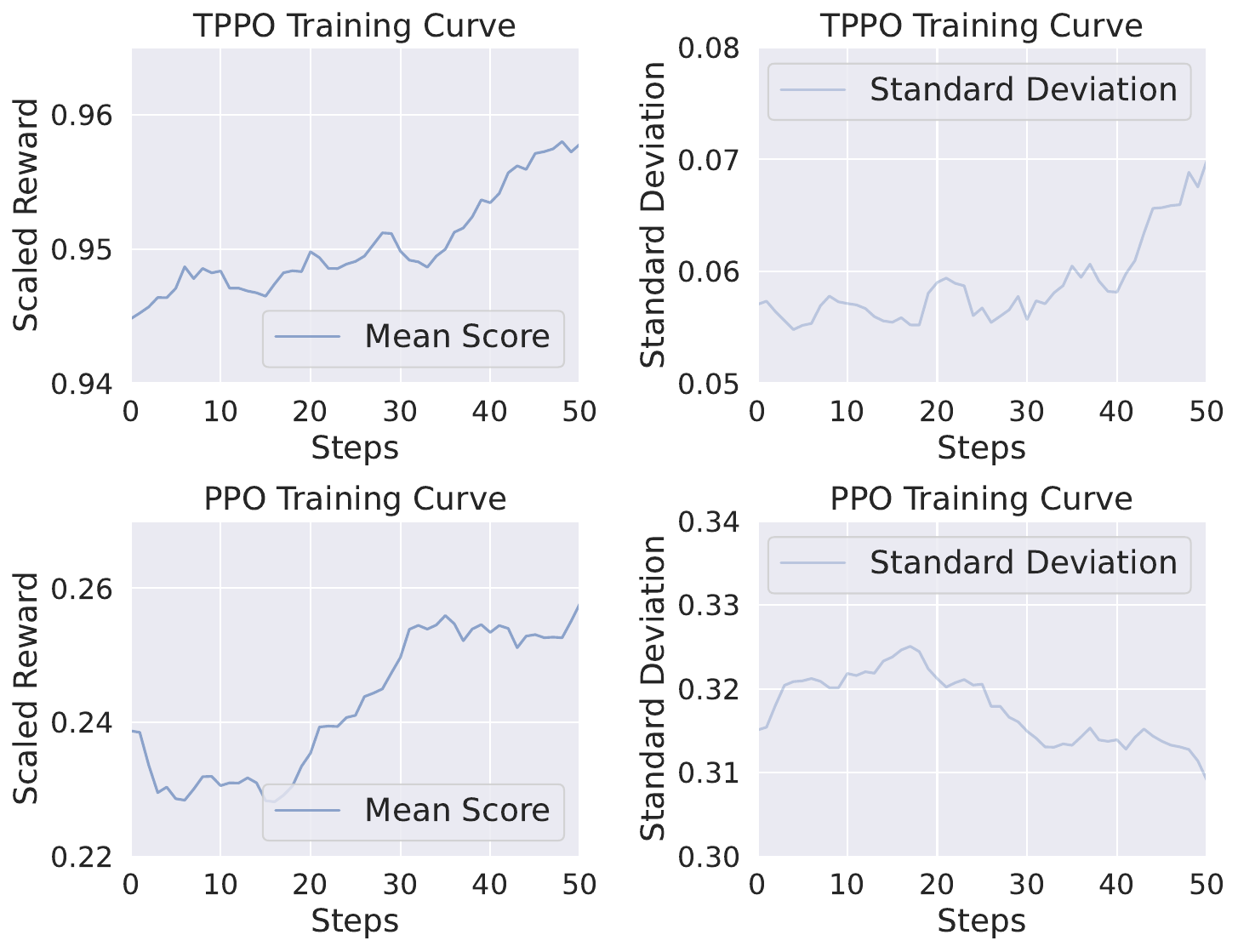}
\caption{PPO Training Curves on Open-source Dataset.}
\label{fig:curve of aol ppo}
\end{figure}

Moreover, we compared the training performance of models obtained using the traditional PPO and our token-level PPO policy on the same training set. As shown in Figure~\ref{fig:curve of aol ppo}, the results indicate that the training of token-level PPO is more stable (smaller variance) and can more easily learn the true preferences of the reward model (faster score improvement). 

\subsubsection{Results of Token-level Reward Model}

\begin{figure}[h!]
\centering
\includegraphics[width=0.5\textwidth]{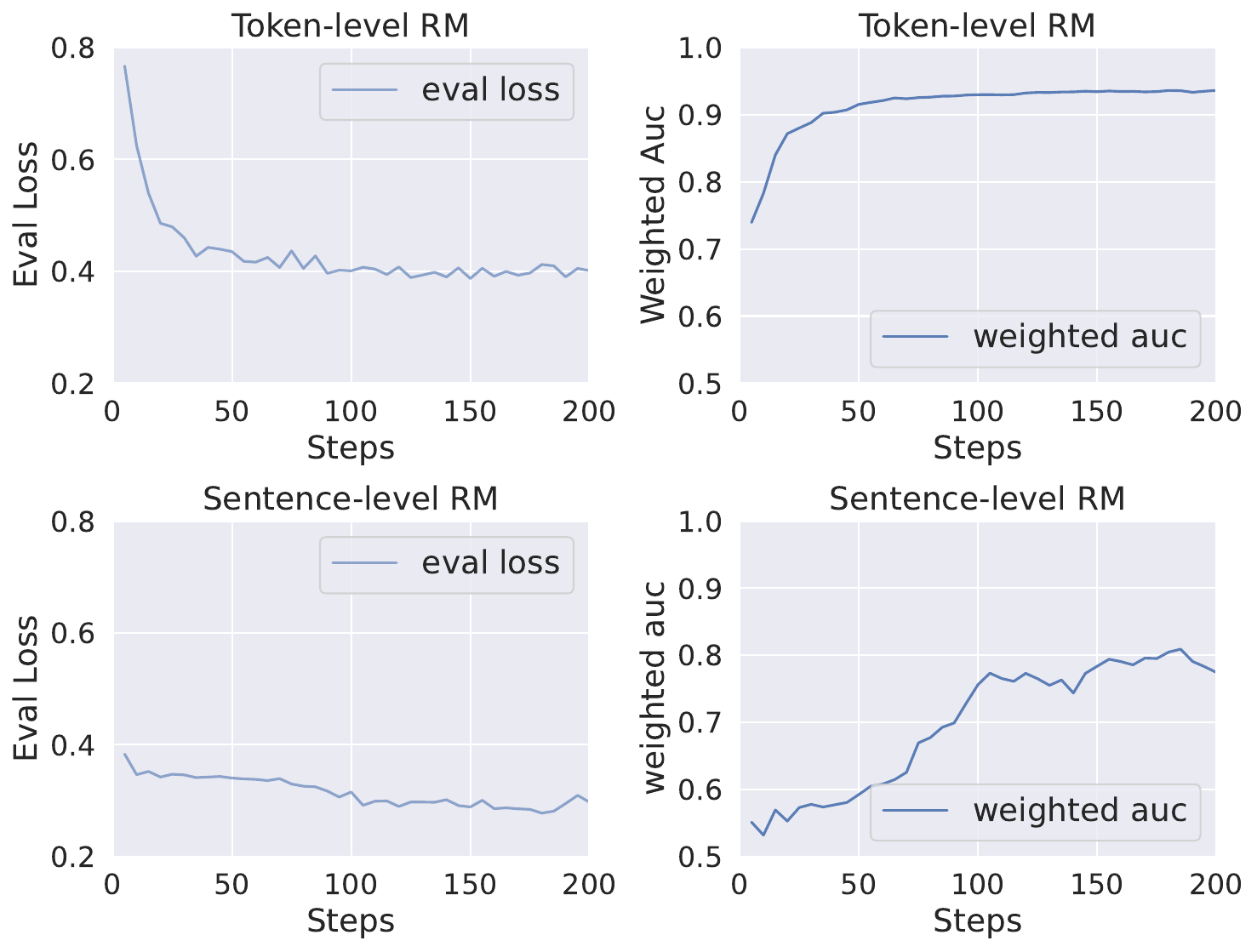}
\caption{Reward Model Training Curve on Open-source Dataset.}
\label{fig:curve of aol rm}
\end{figure}

As shown in Figure~\ref{fig:curve of aol rm}, we compared the training curves of the traditional sentence-level reward model and our token-level reward model, both of which have undergone class balancing to achieve best performance. The comparison of the training curves highlights the advantages of the token-level reward model over the traditional sentence-level approach. Benefited from more granular information, the token-level reward model demonstrates better training stability, faster convergence, and higher performance in terms of AUC~\cite{yang2022aucmaximizationerabig}.

\subsection{Experiments on Industrial data}
\subsubsection{Dataset Description}

As shown in the Table~\ref{tab:dataset_info}, in addition to publicly available datasets, we also collected industrial data from a popular search engine that serves billions of users worldwide. This data includes 400k real user data, from which we filtered 200k data to construct an industrial dataset for the query generation task. In this dataset, the user history consists of 4 components: search, click, purchase, and visit, while the last 10 real search queries in chronological order are used as the target queries to be generated. Furthermore, considering the differences in data distribution, we separately constructed two evaluation sets, eval1 and eval2, each one contains 2k data points collected from different periods.

\begin{table}[tbp]
\caption{Results of Relevance Rate on Open-source Dataset.}
\centering
\begin{tabular}{|c|c|c|c|}
\hline
  & SFT & PPO & TPPO \\
\hline
Relevance rate & 41.25 & 47.65 & 50.00 \\
\hline
\end{tabular}
\label{tab:aol_t1}
\end{table}

\subsubsection{Results of Token-level PPO Policy}

\begin{figure}[t!]
\centering
\includegraphics[width=0.5\textwidth]{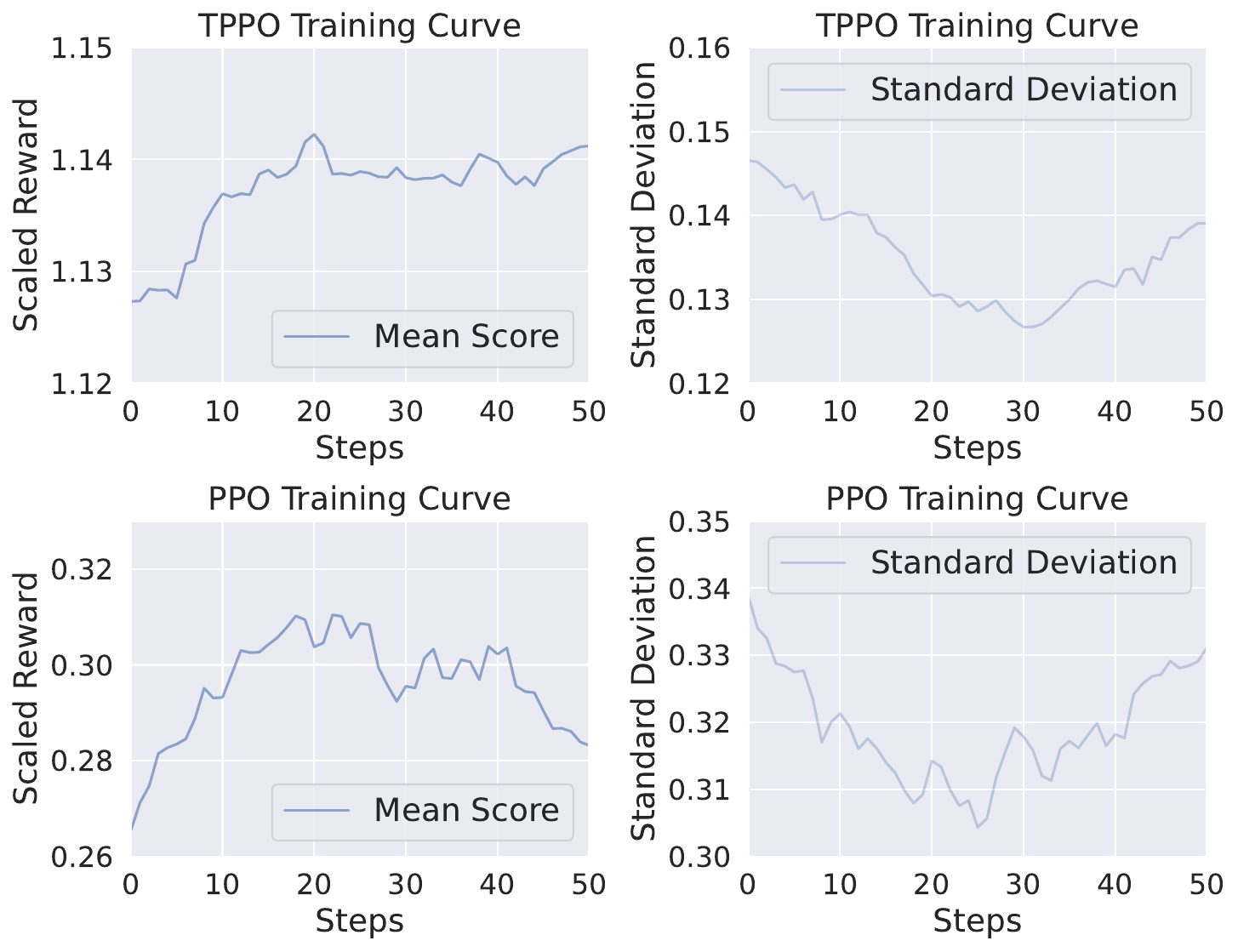}
\caption{PPO Training Curves on Industrial Dataset.}
\label{fig:curve of bingads ppo}
\end{figure}

As shown in Figure~\ref{fig:curve of bingads ppo}, we compared the training performance of models obtained using the traditional PPO and our token-level PPO policy on the same training set. The results indicate that the training of token-level PPO is more stable (smaller variance) and can more easily learn the true preferences of the reward model (faster score improvement). 

\begin{figure*}[t!]
\centering
\includegraphics[width=1.0\textwidth]{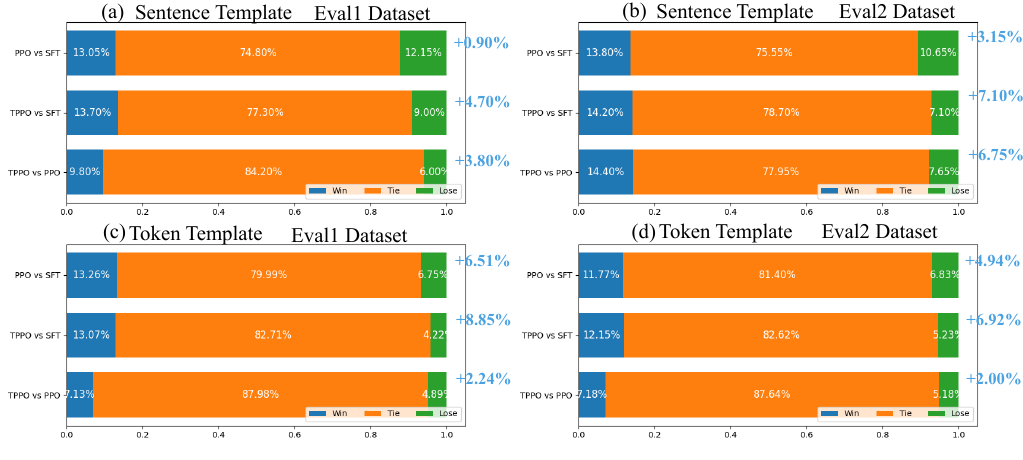}
\caption{Win-Tie-Lose Comparisons on Industrial Dataset.}
\label{fig:gsb_bingads}
\end{figure*}

We set up two scoring templates and two evaluation datasets for the industrial dataset. The sentence template ask Llama3-70B to directly score sentence, while the token template ask Llama3-70B to first score each word and then score sentence. We separately evaluated the experimental results on both scoring templates to verify the consistency and stability of our strategy. Furthermore, considering the timeliness of industrial datasets, we designed two test sets from different months to validate the applicability of our strategy to industrial query generation tasks. Table~\ref{tab:bingads_t1} presents the experimental results on the sentence template, indicating that in the case of directly scoring sentences, our method can effectively improve the relevance rate compared to SFT and PPO. As a complementary, Table~\ref{tab:bingads_t2} shows the experimental results on the token template, demonstrating that when scoring tokens first and then scoring sentences, our method can still stably and effectively improve the relevance rate. As illustrated in Figure~\ref{fig:gsb_bingads}, we also make pairwise win-lose comparisons of the relevance scores of models obtained using SFT, PPO, and our TPPO on the two evaluation set and two template. These experimental results suggest that TPPO better fits the preference.

Note that we did not compare with the LLM without fine-tuning. This is because the model without SFT has high failure rate, unstable output, and poor output quality. We note on a smaller dataset that the failure rate of the mistral-7B model without fine-tuning on the query generation task is 8\%, and the relevant rate is only 45.65\%, which are unacceptable for industrial.

\begin{table}[tbp]
\caption{Results of Relevance Rate on Industrial Dataset, Token-level Template for Evaluating.}
\centering
\begin{tabular}{|c|c|c|c|}
\hline
  & SFT & PPO & TPPO \\
\hline
Relevance rate (Eval 1) & 84.10 & 85.05 & 88.85  \\
\hline
Relevance rate (Eval 2) & 84.35 & 87.50 & 91.45  \\
\hline
\end{tabular}
\label{tab:bingads_t1}
\end{table}

\begin{table}[tbp]
\caption{Results of Relevance Rate on Industrial Dataset, Sentence-level Template for Evaluating.}
\centering
\begin{tabular}{|c|c|c|c|c|}
\hline
  & SFT & PPO & TPPO \\
\hline
Relevance rate (Eval 1) & 85.75 & 92.43 & 94.79\\
\hline
Relevance rate (Eval 2) & 87.40 & 92.25 & 94.21 \\
\hline
\end{tabular}
\label{tab:bingads_t2}
\end{table}
In summary, the use of two scoring templates and two evaluation datasets for the industrial dataset showcases the robustness, stability, and adaptability of TPPO. The experimental results confirm that TPPO consistently outperforms SFT and PPO in terms of relevance rate, regardless of the scoring mechanism or the temporal nature of the dataset. This highlights the effectiveness of TPPO in real-world industrial query generation tasks and its ability to better fit user preferences, resulting in improved query generation quality.

\subsubsection{Results of Token-level Reward Model}

\begin{figure}[t!]
\centering
\includegraphics[width=0.5\textwidth]{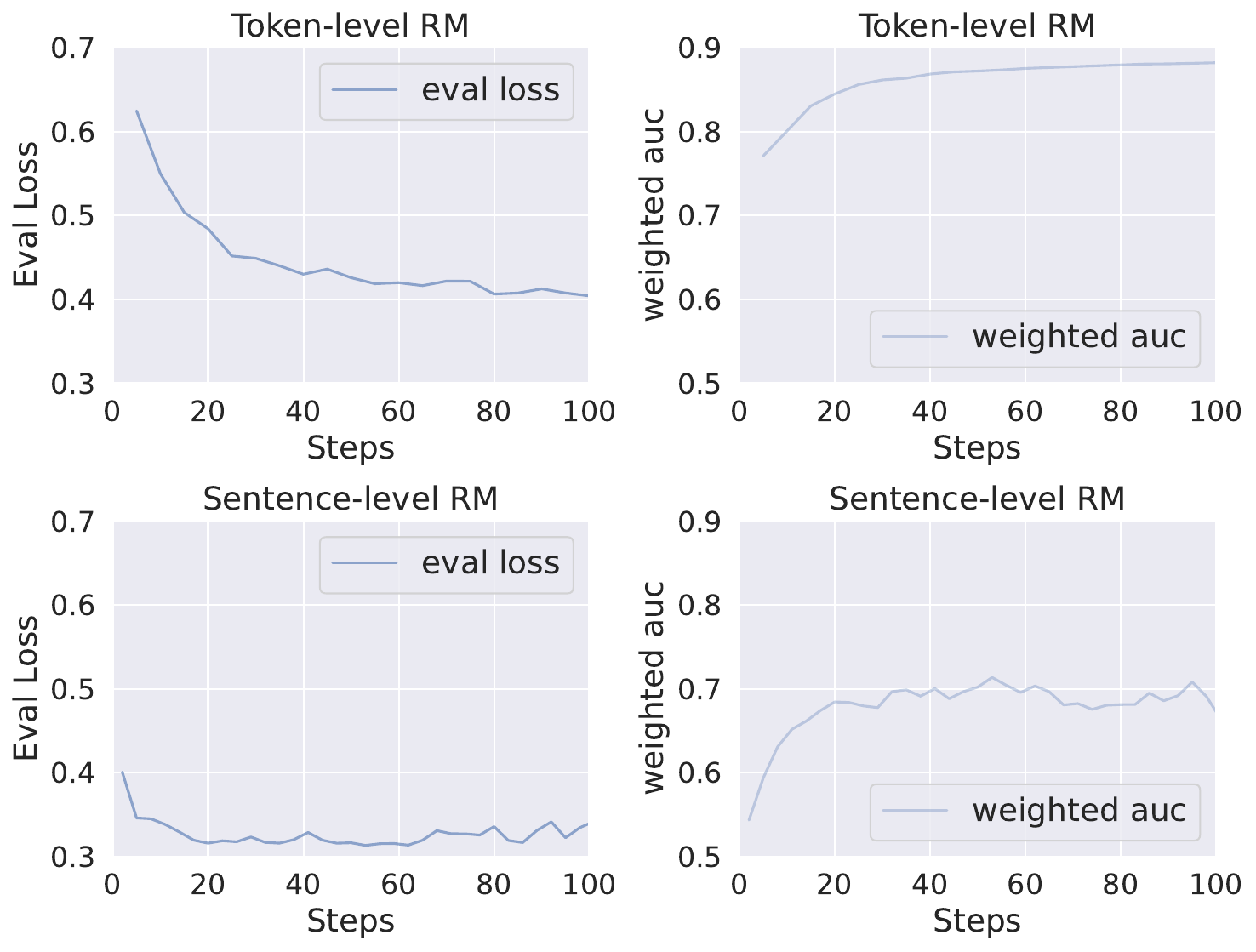}
\caption{Reward Model Training Curve on Industrial Dataset}
\label{fig:curve of bingads rm}
\end{figure}

Figure~\ref{fig:curve of bingads rm} illustrates the superiority of our proposed token-level reward model compared to the conventional sentence-level approach. Both models were trained and evaluated on identical datasets and underwent class balancing for optimal performance. The training curves reveal that by utilizing fine-grained token-level information, our model exhibits enhanced stability during training, achieves convergence more rapidly, and ultimately attains a higher AUC score. These advantages highlight the effectiveness of considering token-level granularity in reward modeling for improved performance and efficiency.

%% file: tex/ablation.tex
\section{Ablation Study}
In the industrial dataset, we conducted ablation experiments for the two core components of the token-level reward model and the token-level PPO policy, respectively. The results highlight the importance of balancing local and global loss in the reward model's learning process and the effectiveness of the length-weighted penalty in constraining the model response length. 

\subsection{Losses of Token-level Reward Model}
\begin{figure}[t!]
\centering
\includegraphics[width=0.5\textwidth]{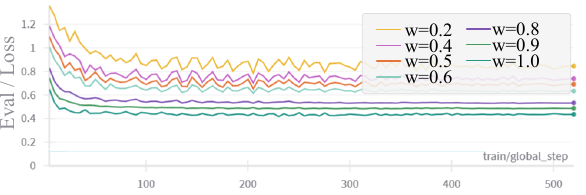}
\caption{Ablation Study of Losses in Token-level Reward Model Training.}
\label{fig:ablation_rm_loss}
\end{figure}

During the training process of the Token-level Reward Model, we used the sum of weighted local loss and global loss as the optimization objective. As shown in Figure~\ref{fig:ablation_rm_loss}, $w$ is the weight of the local loss, and its value is a floating-point number between 0 and 1. Under the same conditions, we plotted the eval loss curves during the training process for different values of $w$. It can be observed that the larger the value of $w$, the smaller the converged loss value, indicating that the local label is a stronger supervision signal compared to the global label. When $w$ is moderate (0.4-0.6), the decrease in loss from the beginning to convergence is the largest, indicating that a balanced combination of local and global loss is optimal for the reward model's learning. This finding highlights the importance of considering both token-level and sentence-level information in the reward modeling process.

\subsection{Length Penalty of Token-level PPO Policy}

\begin{figure}[t!]
\centering
\includegraphics[width=0.5\textwidth]{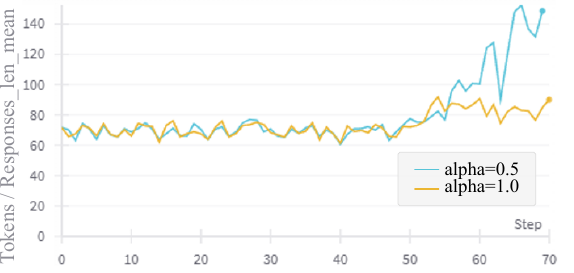}
\caption{Ablation Study of Length Penalty in Token-level PPO Training.}
\label{fig:ablation_lpw}
\end{figure}

In the implementation process of the Token-level PPO Policy, we used the length-weighted penalty controlled by the parameter $\alpha$ to modify the token rewards, in order to achieve length constraints on the model response. As shown in Figure~\ref{fig:ablation_lpw}, under the same conditions, we plotted the change in the number of tokens in the model response during the training process for different values of $\alpha$. 

It can be observed that for $\alpha=0.5$, the constraint is weak, resulting in the issue of excessive output by the model. When alpha is increased to 1.0, the constraint becomes stronger, effectively controlling the number of tokens output by the model. This finding validates the effectiveness of the length penalty in preventing the model from generating overly long responses, which is essential for maintaining the quality and relevance of the generated queries.

%% file: tex/conclusion.tex
\section{Conclusion}

In this work, we proposed Token-level PPO (TPPO), a novel approach that addresses the limitations of PPO in the existing RLHF frameworks for query generation tasks. By introducing token-level reward models and corresponding token-level PPO policies, TPPO mitigates the issues of sparse rewards, mismatch between the traditional PPO in RL and PPO in RLHF, and training instability. Extensive experiments on both public and industrial datasets demonstrate the effectiveness, superiority, and robustness of TPPO compared with baseline methods. TPPO increase the overall relevance rate of generated queries by 2\%-4\% compared with PPO, and has a higher win rate than PPO at 2\%-8\% in item-by-item comparisons. The TPPO training process exhibits better convergence behavior with a stably increasing reward, smaller variance, and improved loss. 

The successful application of TPPO to the query generation task opens up new possibilities for improving the quality and relevance of search results in real-world search engines. Furthermore, the token-level approach introduced in TPPO has the potential to be extended to other natural language processing tasks, revolutionizing the way we generate and refine text. In the future, we aim to develop a more efficient TPPO framework to reduce computational resource requirements. Future research will also explore the application of our token-level to various domains and further demonstrate the generalizability of the techniques introduced in this work.

%% file: tex/appendix.tex
\appendix
\section{Proof of Lemma~\ref{lemma4}}
\label{A_4}

To derive the optimal policy $\pi_{\theta}^*$ for the optimization problem in Eq.~\ref{eq:optimization}, we frame the problem using the method of Lagrange multipliers to incorporate the normalization constraint of the probability distribution $\pi_{\theta}$. The Lagrangian $\mathcal{L}$ is defined as:

\begin{equation}
\begin{split} 
\mathcal{L}(\pi_{\theta}, \lambda(s_t)) = & \sum_{a_t} \pi_{\theta}(a_t|s_t) \left[ A_{\pi_{\mathrm{ref}}}(s_t, a_t) - \beta \log \frac{\pi_{\theta}(a_t|s_t)}{\pi_{\mathrm{ref}}(a_t|s_t)} \right] \\
& - \lambda(s_t) \left( \sum_{a_t} \pi_{\theta}(a_t|s_t) - 1 \right),
\end{split} 
\end{equation}

where $\lambda(s_t)$ is the Lagrange multiplier ensuring that $\pi_{\theta}$ sums to 1 over all actions $a_t$.

Taking the derivative of $\mathcal{L}$ with respect to $\pi_{\theta}(a_t|s_t)$ and setting it to zero gives:

\begin{equation}
\frac{\partial \mathcal{L}}{\partial \pi_{\theta}(a_t|s_t)} = A_{\pi_{\mathrm{ref}}}(s_t, a_t) - \beta \left( 1 + \log \frac{\pi_{\theta}(a_t|s_t)}{\pi_{\mathrm{ref}}(a_t|s_t)} \right) - \lambda(s_t) = 0.
\end{equation}

Solving for $\pi_{\theta}(a_t|s_t)$:

\begin{align}
\beta \log \frac{\pi_{\theta}(a_t|s_t)}{\pi_{\mathrm{ref}}(a_t|s_t)} & = A_{\pi_{\mathrm{ref}}}(s_t, a_t) - \beta - \lambda(s_t), \\
\log \frac{\pi_{\theta}(a_t|s_t)}{\pi_{\mathrm{ref}}(a_t|s_t)} & = \frac{1}{\beta} A_{\pi_{\mathrm{ref}}}(s_t, a_t) - \frac{\lambda(s_t)}{\beta} - 1, \\
\pi_{\theta}(a_t|s_t) & = \pi_{\mathrm{ref}}(a_t|s_t) \exp\left( \frac{1}{\beta} A_{\pi_{\mathrm{ref}}}(s_t, a_t) - \frac{\lambda(s_t)}{\beta} - 1 \right).
\end{align}

The terms $-\frac{\lambda(s_t)}{\beta} - 1$ are constants with respect to $a_t$ for a given $s_t$ and ensure that $\pi_{\theta}$ is a valid probability distribution. They can be absorbed into the partition function $Z(s_t; \beta)$. Thus, we can write:

\begin{equation}
\pi_{\theta}^*(a_t|s_t) = \frac{\pi_{\mathrm{ref}}(a_t|s_t) \exp\left( \frac{1}{\beta} A_{\pi_{\mathrm{ref}}}(s_t, a_t) \right) }{ Z(s_t; \beta) },
\end{equation}

where the partition function $Z(s_t; \beta)$ is defined as:

\begin{equation}
Z(s_t; \beta) = \sum_{a_t} \pi_{\mathrm{ref}}(a_t|s_t) \exp\left( \frac{1}{\beta} A_{\pi_{\mathrm{ref}}}(s_t, a_t) \right).
\end{equation}

This completes the proof.

\section{Prompt Template for Labeling Relevance Score}
\label{template}

\begin{figure}[h!]
\centering
\includegraphics[width=0.5\textwidth]{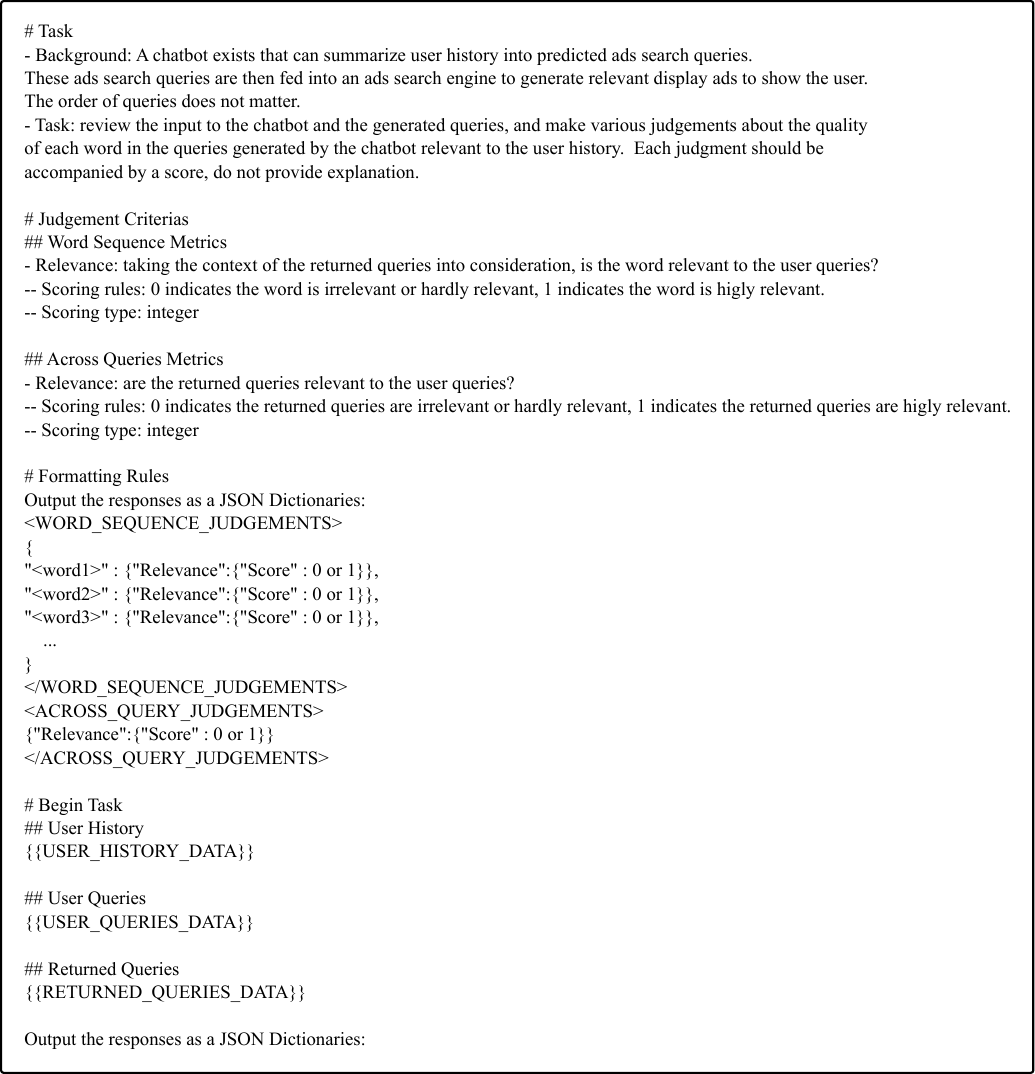}
\caption{Prompt Template for Labeling Relevance Score.}
\label{fig:template_token}
\end{figure}

Fig.~\ref{fig:template_token} shows the template for labeling relevance score.

